\documentclass[runningheads]{llncs}
\usepackage{booktabs}
\usepackage{subcaption}
\usepackage[T1]{fontenc}
\usepackage{graphicx}
\begin{document}
\title{Interpreting the structure of multi-object representations in vision encoders}
%
%
\author{Tarun Khajuria\orcidID{0000-0002-7089-659X} \and
Braian Olmiro Dias \orcidID{0009-0000-1170-7464} \and
Marharyta Domnich\orcidID{0000-0001-5414-6089}\and
Jaan Aru\orcidID{0000-0003-3927-452X}}
\authorrunning{Khajuria et al.}
%
\institute{Institute of Computer Science, University of Tartu, Tartu, Estonia
\email{tarun.khajuria@ut.ee}}
\maketitle              
\begin{abstract}
In this work, we interpret the representations of multi-object scenes in vision encoders through the lens of structured representations. Structured representations allow modeling of individual objects distinctly and their flexible use based on the task context for both scene-level and object-specific tasks. These capabilities play a central role in human reasoning and generalization, allowing us to abstract away irrelevant details and focus on relevant information in a compact and usable form. We define structured representations as those that adhere to two specific properties: binding specific object information into discrete representation units and segregating object representations into separate sets of tokens to minimize cross-object entanglement. 
Based on these properties, we evaluated and compared image encoders pre-trained on classification (ViT), large vision-language models (CLIP, BLIP, FLAVA), and self-supervised methods (DINO, DINOv2). We examine the token representations by creating object-decoding tasks that measure the ability of specific tokens to capture individual objects in multi-object scenes from the COCO dataset. This analysis provides insights into how object-wise representations are distributed across tokens and layers within these vision encoders. Our findings highlight significant differences in the representation of objects depending on their relevance to the pre-training objective, with this effect particularly pronounced in the CLS token (often used for downstream tasks). Meanwhile, networks and layers that exhibit more structured representations retain better information about individual objects. To guide practical applications, we propose formal measures to quantify the two properties of structured representations, aiding in selecting and adapting vision encoders for downstream tasks. Overall, we aim to advance the understanding of object-wise structured representations in vision encoders, thus enhancing their transparency and interpretability. By clarifying how these models bind and segregate object-level information, we enable better-informed decisions for optimal downstream task adaptation, ultimately aligning their behaviour more closely with human reasoning.

\end{abstract}
\section{Introduction}
Vision encoders are foundation models for complex visual inputs in AI, including applications where decisions must be transparent and justifiable, such as healthcare, autonomous vehicles, etc. However, the opacity of these models limits their trustworthiness and applicability in real-world scenarios \cite{lipton2018mythos}. Recognizing the necessity of increasing direct transparency of vision encoders that cannot be guaranteed by post-hoc methods \cite{rudin2019stop}, it becomes essential to focus on the structural representations within the vision encoders themselves. Vision encoders are trained under various objective functions to learn suitable representations of visual inputs, and a \textit{good representation} is the one that contains the correct details for its downstream task \cite{bengio2012representation}. Humans flexibly use structured mental representations for reasoning, where the exact aspect of the representation is generated according to the task context \cite{radulescu2021human,lake2023human}. For instance, in a multi-object scene, we usually focus on only the main objects and infer their relationship to make inferences about the overall situation. However, we also have the ability to focus on individual objects in the scene if the task requires such richer representation abstracted from the scene; e.g., questions about the colour of a chair should be answered irrespective of what scene or surrounding one sees the chair. 

In computer vision, especially for practical applications like robotics, vision encoders trained on large datasets are also expected to be useful for both the scene and the object-level encoding \cite{shridhar2021cliport}. To understand to what extent the representations of pre-trained vision encoders allow for such ability, this study investigates if structured object-wise representations exist in the token space of vision encoders. As previous works \cite{greff2020binding,lovering2022unit,pavlick2023symbols} discuss how such abilities are dependent on having structured representations in a representation system, we build upon this notion of symbols and binding of representation described in these studies to propose two properties specific for an image encoder's representation to be more structured in token space: 
\begin{enumerate}
\item ($M_1$) The model should be able to \textbf{bind} information specific to objects in the image into specific representation units, i.e. a fewer number of tokens can represent the object better than its input token representation; 
\item ($M_2$) The model \textbf{segregates} representation of various objects of the input image in separate sets of tokens, i.e., there should be object-wise information disentanglement in the token space.
\end{enumerate}

We expected these two properties for structured representation to promote better generalisation for the following reasons. First, binding object information into discrete units allows beneficial representation properties to be fully available for novel scenarios in solving downstream tasks. Second, disentanglement of these representation units ensures that one object's presence in the scene does not affect others' representation. This is important for extreme cases where the objects are present out-of-the-scene statistics of the trained distribution. Such spurious correlations in representations of object presence may cause failure for the downstream task.

As we analyse the spatial representations in the vision transformers' token space, our primary questions testing the above properties are the following: \textit{Do transformers represent and maintain object-wise representations?}\textit{ Are these representations disentangled}, i.e. does a particular set of tokens only represent a particular object? We use the COCO dataset \cite{lin2014microsoft} with its instance object masks to characterise the token representations in relation to their input patch information. We set up decoding experiments in a two-object setting (at a time, within a multi-object scene) to determine how the encoders manage the representations of the two objects. We test encoders of three VLMs (CLIP \cite{radford2021learning}, BLIP \cite{li2022blip} and FLAVA \cite{singh2022flava}) and compare them with the larger versions (CLIP-L, BLIP-L), and also check the representations against a CLIP (Resnet X4) with a CNN backbone. We further compare against other pre-training objectives such as VIT (trained for image classification), DINO \cite{caron2021emerging} and DINOv2 \cite{oquab2023dinov2} (encoders pre-trained by image self-supervision). 

To systematically evaluate these two properties, we design complementary tasks: (1) a paired-object decoding setup focusing on two objects at a time from the scene, and (2) a broader 20-object decoding task testing generalization to unrelated categories. In this way, we validate the inferences made by the original task and illustrate the usefulness of our proposed measure to predict the general object encoding ability of the structured representations. 

In the literature, some experimental results were given alongside the release of the original models, which already provide partial inferences about their representations. For example, in CLIP \cite{radford2021learning}, an experiment shows how learning linear decoder on the CLS token representation performs better than their proposed zero-shot classification. In DINO \cite{caron2021emerging}, we can see how different attention heads from the CLS token attend to prominent objects in the scene. While in DINOv2 \cite{oquab2023dinov2}, each token from the last layer can be classified using a linear layer to perform semantic segmentation to some satisfactory detail, showing the object-wise information expressed in the token space of the last layer.

 In our analysis, motivated by these prior indications, we conduct a more systematic investigation into how object-specific representations arise across various encoders. We also test their generality, aiming to provide a more comprehensive understanding of the structure of information expressed in the token space of vision transformers that explains such observations. Our results show that the CLS tokens best models image-level information but, as an effect, prefer to represent the main objects in the scene. Still, object-specific areas in the tokens of the higher layers model hold the most discriminating features about the individual objects. Yet, the object-wise token representations are not disentangled since they can decode other objects in the scene with accuracy far above random guesses. We observe that the VLMs trained on objectives requiring the modeling of multiple objects have better multi-object representations in CLS token than VIT encoders trained for image classification. 
On the other hand, all VIT-based encoders have less segregated and more entangled object representations than CNN-based CLIP (Resnet X4). We highlight the CLS token's inability to model background objects in all models (except CLIP-L) as a possible failure mode when using this token for downstream tasks on multi-object images. Moreover, we show that our two measures of structured representations in the token space of the networks correlate highly with the retention of object-specific information for (less important) background objects. Hence, it illustrates the usefulness of these structured representations in representing multi-object scenarios in the token space and provides a measure to make decisions about appropriate layers to train decoders or adaptors for an appropriate downstream task. 

Our main contributions are:
\begin{enumerate}
    \item We identify two properties to evaluate structured representations in the token space of image encoders. We design experiments and create a relevant dataset using COCO's multi-object images to evaluate these properties and propose a measure related to each property.
    \item Our method lets us understand the nature of information encoding through the lens of structured representations in vision encoders in the token space (about the encoded objects) at different encoder layers. 
    \item  The results from this analysis led us to identify a failure in generalising over multiple objects in the scene. Further, our proposed measures of structured representation ($M_1, M_2$) help against the problem by indicating which tokes/layers and the kind of models can be used to obtain the best representations for the (less represented) background objects. 
\end{enumerate}

\section{Related Work}
Our investigation into structured representations in vision encoders relates to a greater problem of any model's ability to generalise across different scenarios. This overall problem with generalisation in deep learning (DL) methods has been comprehensively discussed \cite{greff2020binding} in light of the need for explicit inductive biases that allow for discrete yet flexible information binding. Discrete symbolic representations are considered a prerequisite for robust compositionality and reasoning \cite{pavlick2023symbols}, and new studies propose that such discrete information binding may originate from training existing models on large datasets \cite{pavlick2023symbols,lovering2022unit}. Beyond discrete symbolic approaches, this issue carries over to the field of meta-learning with its aim to introduce and compositionally utilise modularity in neural networks to improve generalisation \cite{lake2023human}. Concept bottleneck models are a series of models that try to explicitly learn human-interpretable intermediate outputs to be composed into final output labels \cite{de2018clinically,yi2018neural}. However, this interpretability usually comes at the cost of downstream task performance. \cite{koh2020concept} introduced an extra loss in the intermediate layer on the networks for units to align to the interpretable concepts while preserving task performance. Many other works do not bind representations to any explicit intermediate concept but still have explicit discrete bottlenecks in their networks \cite{oord2017neural,locatello2020object}. It has been shown that these networks learn better representations, which further help generalisation on downstream tasks \cite{oord2017neural,locatello2020object,trauble2023discrete}. 

\subsection{Evaluating a trained model's capabilities}
In the case of evaluation of pre-existing networks, many works do not try to estimate the mechanisms or representation structure that lead to particular performance of networks on tasks but rather propose benchmarks to estimate the reasoning capabilities of the ANN models. \cite{antol2015vqa} proposed a visual question-answering benchmark to evaluate the reasoning skills on images, with open-ended and free-form questions and expected solutions in free-form natural language. Visual genome \cite{krishna2017visual} provided content-rich images with explicit annotation of object position and relationships, promoting models that exploit such information. Datasets such as CLEVR \cite{johnson2017clevr}  were designed to correct for biases the models would utilise in the existing benchmarks to perform well without explicit reasoning. The ARO (Attribute, Relation, Order) dataset \cite{yuksekgonul2022and} has been proposed to test the correct binding of information when compositionally representing multi-object scenarios. Particularly in VLMs such as BLIP and CLIP, they find that the model outputs do not bind the compositional properties well and attribute the wrong order and features to objects when describing them. Though these benchmarks reveal model shortcomings at the final output, they do not clarify how (or if) the internal representations disentangle multiple objects.

\subsection{Interpreting model's representations}
In terms of analysis of model representations for concepts, early work by \cite{alain2016understanding} showed the layer-wise progression of decoding accuracy for concepts in CNNs. \cite{cordonnier2019relationship} reveals the inner workings of transformers by showing how trained transformers can implement convolutions and, in the initial layers, form grid-like local attention patterns like a convolution filter. \cite{raghu2021vision} shows that vision transformers differ from CNNs because of their ability to encode both local and global information in the initial layer. In contrast, CNNs exhibit a multi-scale feature representation hierarchy going from lower-scale local information captured in the initial layers to higher-scale global information captured in the higher layers \cite{bronstein2021geometric}. 

Particularly for vision-language models, \cite{cao2020behind} analysed various models trained using cross-model attention, providing insights into the attention patterns between the two modality streams and the relative contribution of each modality towards downstream tasks. Further, they found function-specific attention heads in the pre-trained models. VL Interpret \cite{aflalo2022vl} was designed as a visualisation tool to interpret the vision-language model's instance-specific and aggregate statistics over attention distribution. Further, the tool helps visualise token representation as it passes through various network layers.

In contrast to these generic visualisation tools that look into the model’s functioning or attention-based approaches, other studies inspect model representations, evaluating a specific computational functionality. \cite{lepori2023break} finds modular sub-networks in trained ANN models functionally responsible for separate tasks. Multiple studies further look into the notion of concepts in trained vision-language model representations. A study by \cite{yun2022vision} designed a test to check if primitive concepts emerge in the network’s representations, which are used compositionally for downstream tasks. \cite{lovering2022unit} defines tests for concepts in visual representations according to Fodor’s criteria \cite{fodor1998concepts} and tests these criteria using a controlled synthetic dataset. Building on these insights, our work formulates explicit hypothesis about structured representations regarding object-wise binding and disentanglement in a well-defined scope of layer-wise distribution of representations in visual encoders. We explore natural scenes rather than heavily controlled settings, thus complementing the more synthetic or attention-centered studies above.

\section{Methods}
\begin{figure*}
  \centering
   \includegraphics[width=1\linewidth]{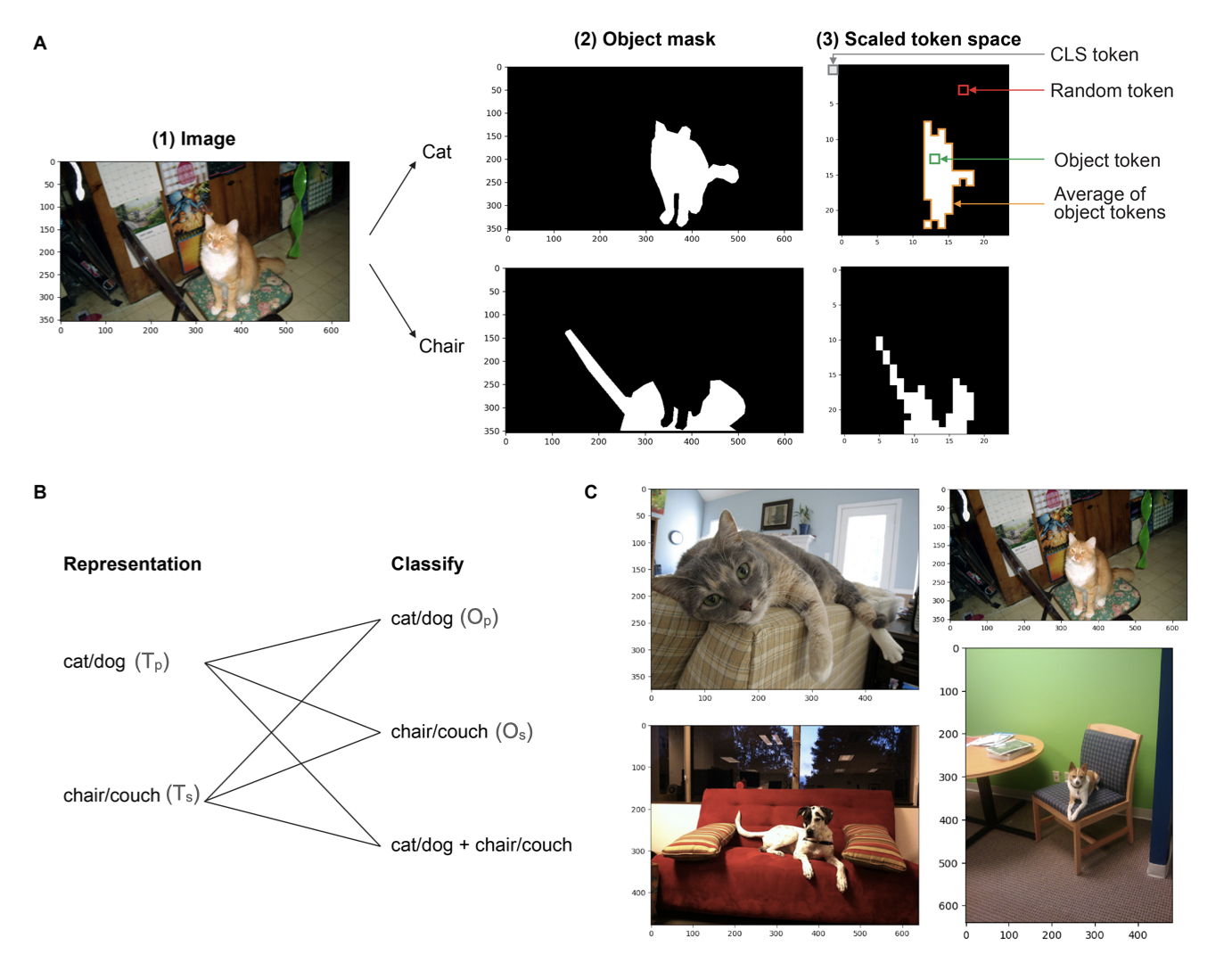}

   \caption{\textbf{A.} Explanation of how the token representations are obtained. We analyse four kinds of tokens in this study: 1) CLS token: The special token usually used in models for downstream tasks; 2) Avg\_obj(Object-specific token): obtained by averaging the token representations of the object-masked tokens as shown in the figure. 3) Random\_obj (Object-specific token): Rather than averaging, we sample one of the tokens from the masked token space of the object 4) Random: Obtained by sampling any random token from the token space other than the CLS token
    \textbf{B.} Describes the experimental setup in which we perform decoding in paired object tasks; each object-specific representation decodes 1) the object itself, 2) the other object in the image, and 3) the combination of both objects.
    \textbf{C.} Shows a sample paired object decoding task; given an image, the task is to decode if it contains object1 (cat/dog), object2 (chair/couch) or a combination of both.}
   \label{fig:methods}
\end{figure*}
We train probes and use similarity measures to evaluate the network's representational structure. In the following section, we formally define measures related to our proposed properties of structured representations and describe the details of the experimental setup, data, and the networks analysed.

\subsection{Formal Definition}

 Our method quantifies the information structure in layers of an image encoder by measuring two properties. Given an image \(I\), containing at least two objects, denoted as a primary object \(O_p\) and a secondary object \(O_s\), we denote the tokens originating from the primary object as \(T_p\) and those originating from secondary objects as \(T_s\). Across a test set, the accuracy of a probe trained and evaluated using representation from \(T_p\) to decode object \(O_s\) is given by \(A_{ps}\). Similarly, \(A_{pp}\),\(A_{sp}\) and \(A_{ss}\) refer to the other token-object combinations. 
 
 In this study, for the first property (binding), we measure the ability of a localised region to decode the individual object. More precisely, we focus on tokens \(T_s\) 's ability to decode object \(O_s\) linearly. Hence, we measure this over a set of images by \(A_{ss}\). 
 Thus, $$M_1 =A_{ss}$$
 The second property (segregation) estimates the mixing of information between the tokens from other objects. We estimate if representations from tokens \(T_s\) can decode object \(O_p\). Over a set of images, we use the decoding accuracy \(A_{sp}\) to estimate this entanglement between token representations, which is the opposite of segregation. For better comparison among layers, we scale this value by the decoding accuracy of that object for from own token representation, i.e. \(A_{pp}\). Hence, the final measure for entanglement of information is given by \(\frac{A_{sp}}{A_{pp}}\). $$M_2 = \frac{A_{sp}}{A_{pp}}$$
 
\subsection{Experimental Setup: Paired Object Task}

The main experiment for assessing how token space encodes multi-object information is the \textit{paired object task}. The objective of this experimental setup is to have images that contain only one of the two primary objects and one of the four secondary objects. The basic inference we make is that \textit{given a particular token representation of the image from the encoder, can we reliably infer both the primary and the secondary object?} 

We try to infer the multi-object relations in a pairwise manner as the objects vary in location, size and occur in different scene context. We select a total of 6 object sets that help generalise over different object types. The particular importance of the objects is selected by their relative importance to the scene, which is further formalised in later experiments by quantifying their mention in the COCO captions for the particular image.     
We decode the combination of objects in the image from a single token or an average of tokens obtained from various parts of the image (see Fig. \ref{fig:methods}.A). The tokens we are interested in include are: 
\begin{enumerate}
\item CLS token: the token used and trained for the encoder's downstream task, 
\item Average token (avg\_obj): the average of token representation obtained from the object, 
\item Random object token (random\_obj): a single token randomly sampled from the object, 
\item Random token (Random): a randomly sampled token from the image that served as a baseline.
\end{enumerate}

The token representation is obtained at the output of each layer. To identify the tokens originating from an object, we scale the segmentation mask of the object to the size of the token space. 

The paired-object probing task is designed as a classification task with the following settings: we use the tokens originating from 1) Primary object (\(T_p\)), 2) Secondary object  (\(T_s\)) and use it to decode 1) Primary object \((O_p\)) category, 2) Secondary object (\(O_s\)) category, 3) Combination of primary and secondary object categories (see Fig. \ref{fig:methods}.B). We train and evaluate the probes with a train/val/test setup with 80/10/10 percent data splits. The reported accuracies are all final test set accuracies. All probes are linear and use the Scikit-learn \cite{pedregosa2011scikit}'s perceptron implementation, with its default parameters.

\subsection{Experimental Setup: Global Probe}
The global decoding task is proposed to test the generalisation of the numbers obtained by the paired-object task on completely different categories of objects. It also allows us to verify using a more complex multi-class (20 class) probe that the trends in the paired object task are not a result of a simple 2 class or 4 class decoding (as in the paired-object task). Finally, in this analysis, we formalise another notion of object importance in the test set data split between objects mentioned in the caption as 'in the caption' and objects not mentioned in the caption as 'not in the caption'. This further helps test the difference between main and secondary objects apart from the category level classification as primary and secondary objects used in the paired object task. 

In this task, object-specific probes are trained, i.e. using Random\_Obj and Avg\_obj tokens to decode the object category from which those tokens are taken. These probes utilise the first 40000 images from the MSCOCO train set to train a probe for the layer and each token type. The 5000 images in the validation set are used for testing. As we are testing for generalisation, we exclude the object categories used in the six object sets used to calculate our structured representation measures given in Table \ref{tab:dataset} and randomly choose 20 object categories (given in Global section of Table \ref{tab:dataset}) in the COCO dataset for this evaluation.

\subsection{Dataset}
We needed instance segmentation masks to associate the tokens with the object in the image. Hence, we used the COCO dataset and created subsets combining object categories. We then excluded images with more than one object category of primary or secondary type. We also sub-sample to balance the co-occurrence of all object combinations between the primary and secondary categories. This balancing step is important to measure entanglement correctly, as otherwise, the difference in the co-occurrence of primary and secondary objects allows the decoder to learn the presence of one from the other. While choosing the objects across primary and secondary class we preferred objects more likely to interact in the scene. Within the primary and secondary categories, we preferred objects that are similar to each other, so their embedded representations are not naturally distinct, increasing our probe's sensitivity. We obtain six task sets with a total of 16,288 images. For example, the first task contains a combination of objects from two sets, i.e. a primary object: animal (cat/dog) and a secondary object: furniture (chair/bench/bed/couch). The dataset and its object sets are detailed in Table \ref{tab:dataset}. We call this set of tasks 'object pair decoding tasks'.

For the generalisation task, global probes were trained and tested on a larger dataset of 20 random classes (given in Table \ref{tab:dataset}) beyond the ones used in the main analysis and measures. We selected the first 40000 images in COCO's training set for training and the 5000 images in the validation set for testing.

\begin{table*}[tb]
  \centering
  \caption{For paired object decoding, we use 6 object sets with different numbers of images in each set. Each set contains images with different variations of objects. For the global object decoding task, we tested generalisation on 20 randomly chosen objects.}
  \begin{tabular}{p{0.7in}p{0.57in}p{0.57in}p{0.65in}p{0.7in}p{0.7in}p{0.6in}}
  \toprule 
 \textbf{Paired-Object Probe} & \textbf{Set 1}   & \textbf{Set 2}               & \textbf{Set 3}    & \textbf{Set 4}      & \textbf{Set 5}         & \textbf{Set 6}    \\
    \midrule
\textbf{\# Images} & 2414    & 5042                & 1953      & 2143       & 938            & 3738           \\
\textbf{Primary Object} & cat, dog & dining table, person  & train, bus & tv, laptop & microwave, oven & motorcycle, car  \\
\textbf{Secondary Object} &
  bench, chair, couch, bed &
  pizza, knife, cup, cake &
  traffic light, bench, backpack, handbag &
  mouse, remote, keyboard, cellphone &
  bottle, spoon, knife, cup &
  traffic light, handbag, backpack, bicycle\\
\midrule
\textbf{Global Probe} & \multicolumn{6}{p{3.8in}}{sheep, bear, banana, potted plant, bowl, toilet, horse, apple, fire, parking meter,
 handbag, snowboard, broccoli, giraffe, stop sign,hydrant, cow, tie, hot dog, truck, wine glass}\\

\bottomrule

\end{tabular}
  
  \label{tab:dataset}
\end{table*}

\subsection{Linear Probing}
 We probe the representations in pre-trained networks for our analysis. Probing an information system involves obtaining a representation, usually in the form of a vector from the system in response to an image. Then, we estimate if that particular vector can classify information about the stimuli correctly. The kind of information the probe can learn to classify, and the complexity of the probe (i.e. is it just a linear classifier or a complex multilayer NN) indicates the nature of information present in that layer’s representation. We do not use a more complex probe because it is difficult to verify if the classification performance for them is a function of better representation or due to the learnt complex relationship in the probe \cite{alain2016understanding,hewitt2019designing}. Hence, we obtain representations from various parts (layers and spatial sections, i.e., tokens) of the network and use them to understand the network by looking at the ability of a linear probe to classify it.

 \subsection{Analysed models and their configuration}
 We analysed the image encoders of the nine pre-trained models. 
\begin{itemize}
    \item \textbf{Vision-Language Models:} BLIP (ViT-B/16 and ViT-L/14) for image captioning, CLIP (ViT-B/16 and ViT-L/14) for image-text matching, and FLAVA with an additional multimodal encoder on top of vision and language encoders.
    \item \textbf{Baseline ViT-B/16:} trained on ImageNet21k for image classification.
    \item \textbf{CNN-based CLIP (ResNet X4):} an alternative architecture compared to ViT-based CLIP.
    \item \textbf{Self-Supervised Methods:} DINO and DINOv2, each trained on images without labels.
\end{itemize}
 In CNN analysis, we obtained feature cells instead of tokens, i.e., we used the vector representation of a cell in the feature map by accumulating all the filter outputs at the cell location. In both transformers and CNNs, the place on the feature map representing the object is computed by scaling the object segmentation map to the size of the feature map at the layer. The images are pre-processed with the standard pre-processing function and setting provided along with the pre-trained network instances.

 \section{Results}

In this section, we will discuss the results and observations made while looking at the vision encoders through the lens of structured representations. In subsection 1) we see how the images from the paired-object task allow us to see an aggregate and instance-specific view of the representation of various objects in different types of tokens and layers. Subsection 2) shows how the object representations are spread across object-specific tokens and CLS tokens with the example of BLIP. 3) We further discuss the results for all networks, discussing the key variations in the distribution of representations based on the type of network and the training objective. 4) Following up on the result of object-specific tokens having better representations than CLS tokens, we illustrate this effect by relating it to downstream tasks. 5) Finally, we correlate the results obtained till now to unrelated objects in the global decoding task. We see how the measures of structured representation can be used to estimate the layer and network better at retaining object-specific representations in multi-object scenes.

\subsection{A snapshot of network's global and image-specific representation}
\begin{figure*}
  \centering
    \begin{subfigure}[t]{0.7\textwidth}
      \centering
      \includegraphics[width=0.90\textwidth]{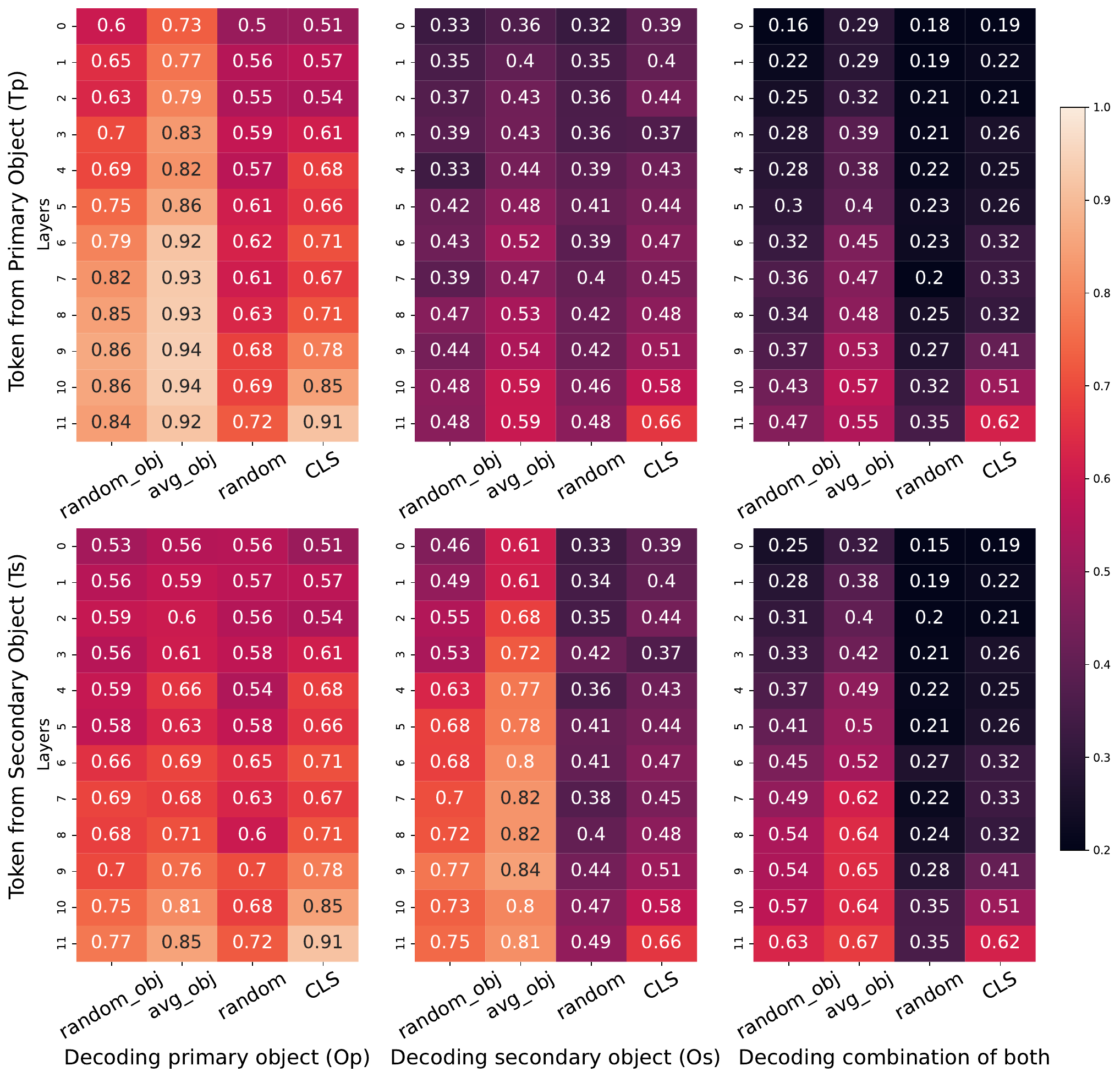}
       \caption*{(a)}
    \end{subfigure}\hfill
    \begin{subfigure}[t]{0.3\textwidth}
      \centering
      \includegraphics[width=0.99\textwidth]{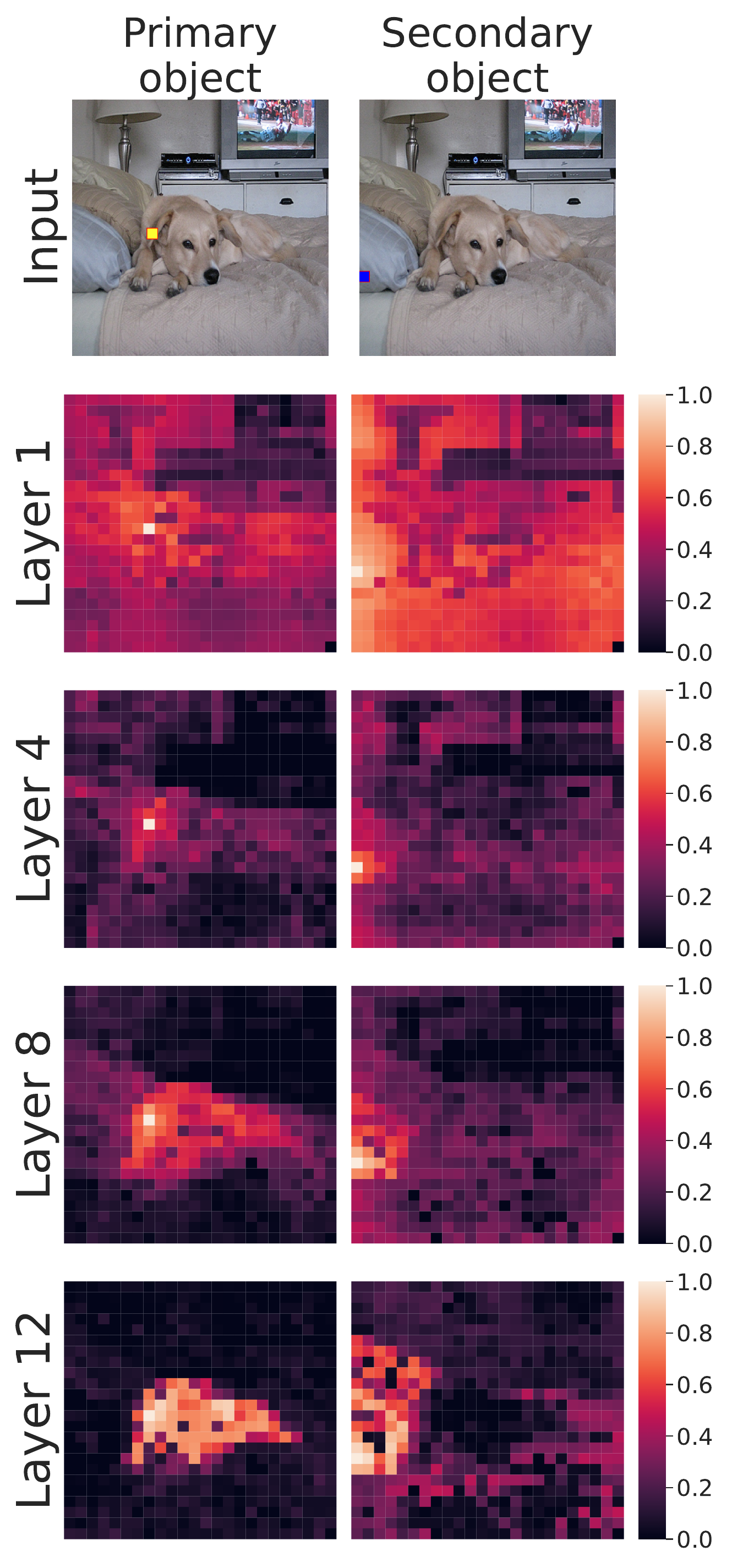}
      \caption*{(b)}
  \end{subfigure}\hfill
   \caption{\textbf{a.} Paired object decoding task results for BLIP across layers: Average decoding performance for different layers (y-axis) and token types (x-axis) over 6 tasks for BLIP. In the subfigures, the y-axis contains variations of where the object-specific tokens (random\_obj and avg\_obj) are obtained. The different columns show results for 1) decoding the primary object 2) the secondary object 3) the combination of both objects in the image. The decoding pattern remains after averaging, with the tokens from the objects modelling the most useful information for categorising the objects. The object-specific tokens are much better than the CLS token, which has to capture the larger scene context. \textbf{b.} Visualisation of cosine similarity of highlighted token to other tokens for a token from primary and secondary objects at various layers of BLIP model.}
   \label{fig:2}
\end{figure*}
Decoding the objects in a multi-object scene using the trained model's representations gives us a picture of how the information about the scene is organised in the network. In our study, we have used two kinds of decoding tasks (paired object and global decoding tasks) to estimate the architecture-level organisation of these representations over a set of images. We also generate representation similarity maps to analyse the representation of a single image in a particular model. 

Using the paired object decoding tasks, in Fig. \ref{fig:2}.a, you can see the detailed results for decoding accuracies for the four types of token representation used in our analysis for the paired object decoding tasks. This provides a more detailed global picture of the model's representations of object pairs from the images. For an instance-wise image-level analysis of representations, we generate similarity maps of the object representations that show how a particular token's relation to other tokens changes across layers in the model for a particular image. Fig. \ref{fig:2}.b shows the representation similarity map (visualising cosine similarity) for all tokens to two tokens (marked with yellow and blue squares in the original images) in an image 1) to the primary object 2) to the secondary object. One can notice how the representations of tokens originating from the same token start having similar representations as we move toward the upper layers. The tokens from the primary object (dog) acquire similar representation, but for the secondary object (pillow), many tokens outside the object also have high cosine similarity in the last layer. 

\begin{figure*}
  \centering
   \includegraphics[width=1\linewidth]{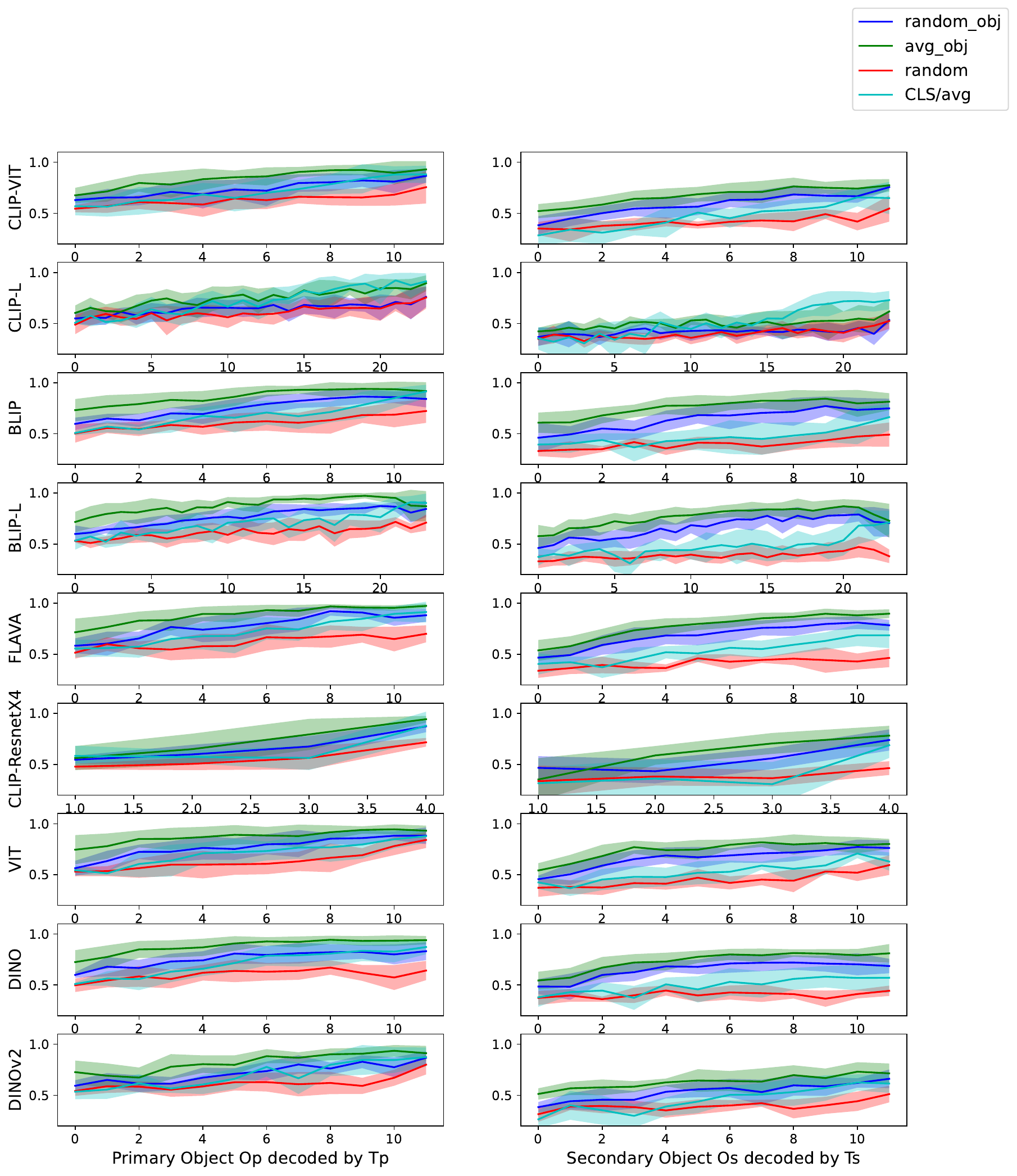}

   \caption{Layer-wise test set decoding accuracy for primary and secondary objects for pre-trained models in the study. Results for all models are shown in the appendix. The accuracies are averaged over the six object sets. In each sub-graph, the y-axis denotes the decoding accuracy, and the x-axis denotes the layer at which the accuracy was observed. We observe consistent decoding trends across models with a few variations reported in Section \ref{sec:Variation across models}.}
   \label{fig:3}
\end{figure*}

Finally, we can see that the objects in the network can be linearly decoded considerably above random accuracy by using either a single token representation originating from the object or by using an average of token representations from the object. The general trend followed by these decoding accuracies can be seen in Fig. \ref{fig:3}. The global object decoding probes, with their 20 class classification setting with larger train and test sets (40000 and 5000 respectively), provide a solid baseline of the ability of the token's representations to decode between multiple objects. This setup also tests on completely different sets of objects compared to the paired-object task (Objects in Table \ref{tab:dataset}). The high correlation of these accuracies for each network (see Fig. \ref{fig:5}) with paired-object classification tasks' results (i.e., $M_1$ as it uses Avg\_Obj accuracy from the paired object decoding task) shows that the accuracies are not obtained due to an easy 2-way or 4-way classification task setup in the paired object tasks, nor by overfitting to a particular set of objects.

\subsection{How different token representations encode objects, their interaction, and their importance }
In the paired-object tasks, the images consist of a primary object and a secondary object. We see the representation of these two object combinations in each image in four types of token representations from the models. Based on the results from the BLIP model shown in Fig \ref{fig:2}, we now discuss the general trend of representations across models. Specific differences are discussed in subsection \ref{sec:Variation across models}. 

 In our results (see Fig. \ref{fig:4}), in all pre-trained models, the primary objects are decoded equally well by the CLS token as the average token representation of the object. There is a decrease in decoding accuracy from primary to secondary object categories, partially due to the added complexity of a 4-way classification for the secondary object.  However, the CLS token decodes the secondary objects with notably lower accuracy than both the object-specific tokens (avg\_obj and random\_obj). The CLS tokens, optimised for the downstream tasks in each network, are expected to model the best information about the scene. Yet this also means that not all objects are linearly decodable by the CLS token. We observe that the object-specific tokens have better object decoding accuracy compared to the CLS token (in BLIP, Fig. \ref{fig:2}, the primary object has CLS: 0.91 vs Avg\_Obj: 0.92 and the secondary object has CLS: 0.66 vs Avg\_Obj: 0.81). There is a particular degradation in decoding performance using the CLS token for secondary objects, where its accuracy is far below the decoding accuracy of object-specific tokens (avg\_obj and random\_obj). 

Previous analyses showed that object-specific tokens have the highest accuracy for decoding the particular object from which they originated. Still, they also have high decoding accuracy for the other objects in the image. This accuracy is far above random guess and allows these tokens to decode a combination of objects in the current image (see Fig. \ref{fig:2}, for BLIP results). For example, the final decoding accuracy for the combination is 0.57, when using the primary (avg\_obj) object token and 0.65 when using the secondary one. In Fig. \ref{fig:2}, one can notice that the primary avg\_obj token is much worse at representing the secondary objects than the other way around. These findings have implications for the disentanglement in object-specific tokens. Further, the CLS token still shows a decent decoding accuracy as it is optimised to capture more global information from the image. However, it remains lower than the object-specific secondary tokens, due to its weaker capturing of information about these secondary objects in the scene (CLS token decoding accuracy for the secondary object is 0.66 compared to 0.91 for the primary object, as seen in Fig.\ref{fig:2}). 

\subsection{Variation of representations across models with different training objectives and architectures: key insights}
\label{sec:Variation across models}

In this section, we compare the representations of the models with each other. Overall, across the models, the decoding accuracy of the object-specific tokens of the paired-object decoding tasks correlates with their global decoding task performance (see Fig. \ref{fig:5}). The relative accuracy trends for various tokens are similar across most VLM models. A notable observation is the higher overall decoding performance of FLAVA, BLIP and BLIP-Large models, which also shows a higher segregation of object-wise representations (i.e., the random token accuracy is fairly lower in the last layer compared to object-specific tokens). These models also score high on our object binding measure and relatively low on entanglement (except BLIP-L and CLIP-L). The two self-supervised models, DINO and DINOv2, show a slightly lower score on object binding for secondary objects, yet their representations are more disentangled, which allows them to model the background objects relatively better. Interestingly, we observe in Fig.\ref{fig:3} that in DINOv2, the decoding accuracy of random\_obj is higher than other models, aligning with its reported performance on use for semantic segmentation tasks using token-level classification. 

 We observe a difference in representations' structure due to architecture (Transformer vs CNN) and specific training on multi-object tasks (i.e., ViT vs other models).  Consequently, there is a significant decrease in decoding accuracy using the random CNN unit representation compared to random ViT tokens (in the last layers, primary object: ViT 0.84 vs CNN 0.72; secondary object: ViT 0.54 vs CNN 0.45; see Fig. \ref{fig:3}). Further, the object-specific tokens in CNNs have lower accuracy while decoding the other objects in the scene than their ViT counterpart (in the last layers, primary object: ViT 0.88 vs CNN 0.8; secondary object: ViT 0.6 vs CNN 0.59). This indicates that CNN has less entanglement of object-specific information across objects than the ViT counterpart in CLIP. We attribute these results to CNNs not having the ability for the information to travel across units in each layer; hence, the object information remains more localised.
 ViT trained on ImageNet21k shows the least differentiation between object-specific (Random\_Obj: 0.88)  and other tokens (Random: 0.84) compared to the other Transformer models. Here, a random token decodes the object with almost similar accuracy as the CLS token (CLS: 0.88; see Fig. \ref{fig:3}), and tokens from one object can similarly decode the category of other objects in the scene. Hence, the scene-level information appears more uniformly dispersed in the representations of ViT tokens trained only on single object classification tasks. 
 We note that this difference can be because the other models have been trained on objectives that require correct modelling and representation of multiple objects for downstream tasks like text matching, captioning, etc. Hence, the differentiation of object-specific tokens from random tokens and, consequently, the segregation of information is more explicit in the last layer in many of these networks than in ViT(trained for single object classification). 

Turning to the \textit{global decoding task}, we see that the newer VLMs like FLAVA and BLIP have higher accuracy for their object-specific representations than counterparts like CLIP (see Fig \ref{fig:5}). A larger model (BLIP-L) performs better than its base version. However, the layer-wise analysis (see Fig.~\ref{fig:3}) indicates that the CLS token outperforms object-specific tokens for CLIP-L. This points to a different binding mechanism, where the object-specific tokens have low classification accuracy across object categories, but the CLS token effectively binds both primary and secondary objects in the higher layers. Consequently, one can note that CLIP-L has the best scene-level decoding accuracy (0.66) when using the CLS token to decode the combination of primary and secondary objects. On the other hand, ViT trained on single object classification has the poorest accuracy in decoding a combination of objects using the CLS token (0.49). However, its object-specific regions maintain fairly good representations for individual objects (Avg\_Obj $A_{pp}$= 0.93 and $A_{ss}$=0.80).

Lastly, the decoding accuracy for object-specific tokens degrades the most on secondary objects, which is also correlated with its lower accuracy on objects 'not mentioned in the caption' in Fig. \ref{fig:5}. Our layer-wise analysis in Fig.\ref{fig:3} also shows that in BLIP-L and many networks like ViT, DINOv2, CLIP, and BLIP, the secondary object accuracy starts decreasing in the final few layers, showing the effect of the training objective. We explore that effect further in the next two sections.  

\subsection{Special token may not be special for your downstream task with multi-object images}

\begin{figure*}
  \centering
    \begin{subfigure}[t]{0.72\textwidth}
      \centering
      \includegraphics[width=0.90\textwidth]{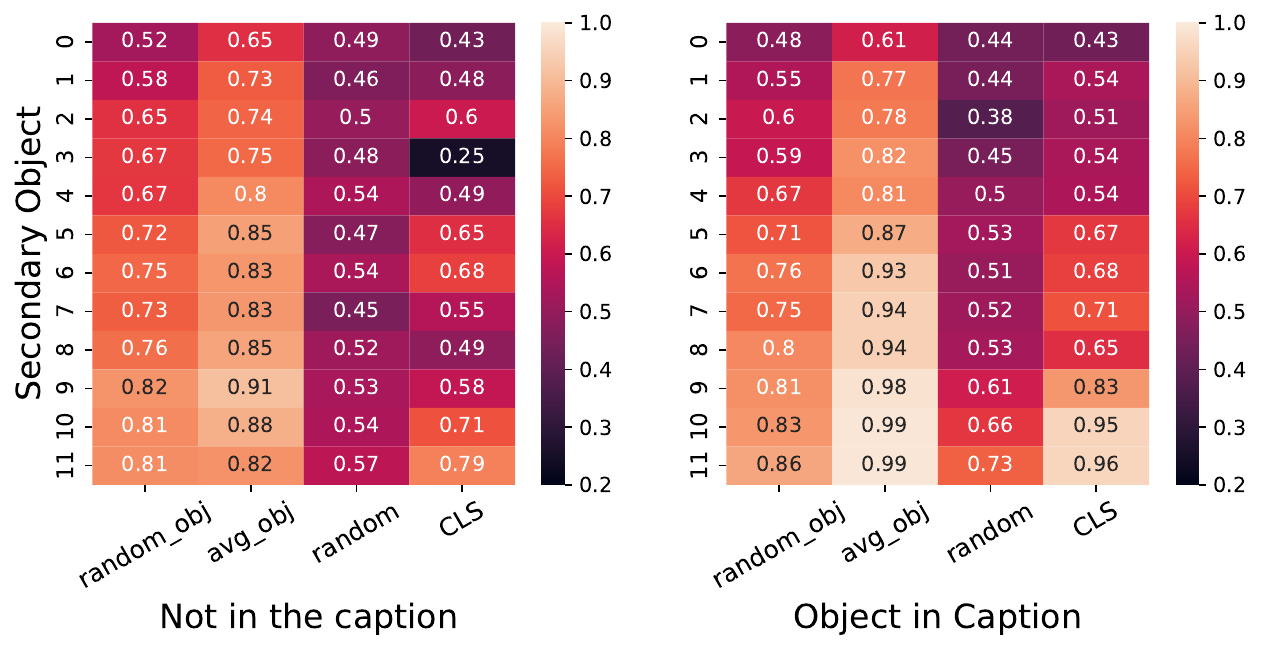}
       \caption*{(a)}
    \end{subfigure}\hfill
    \begin{subfigure}[t]{0.28\textwidth}
      \centering
      \includegraphics[width=0.99\textwidth]{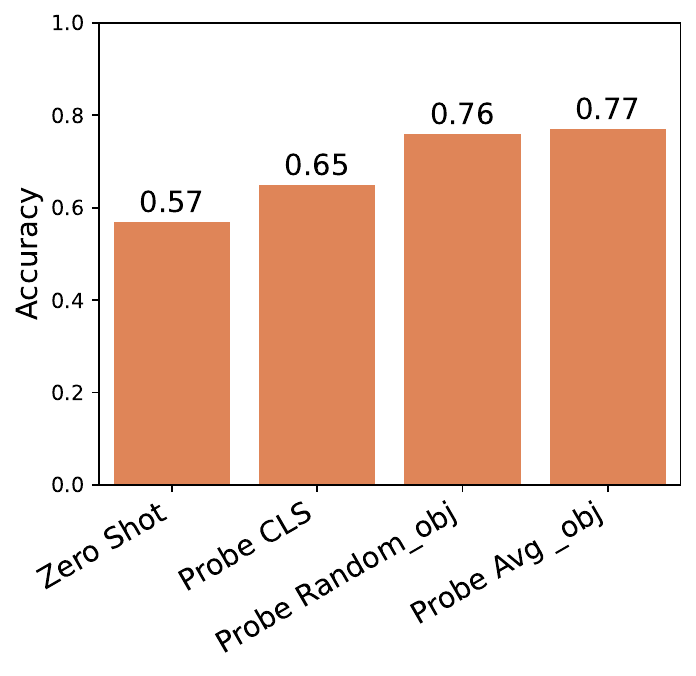}
      \caption*{(b)}
  \end{subfigure}\hfill
   \caption{\textbf{a.} Variation in decoding accuracy between instances of objects ‘in caption’ and ‘not in caption’. Each subplot represents the decoding of the object by its object-specific representation. \textbf{b.} Object detection accuracy on the secondary objects in paired using probes trained on three different token representations from CLIP and zero-shot CLIP accuracy.}
   \label{fig:4}
\end{figure*}

Following up on the observations about the lower decoding accuracy of secondary objects for CLS tokens, we wanted to check if the low accuracy results are related to the object’s importance in the downstream task (captioning in the case of BLIP). Hence, we analysed the decoding accuracy for two sets of data. We divided the data based on whether the object was directly mentioned in the caption generated by the model. Due to this split, we rejected object sets with less than 400 samples in either of the new sets. Therefore, we are left with 3 object sets whose average results are reported in Fig. \ref{fig:4}.a. From Fig. \ref{fig:4}.a, one can see a decrease in decoding accuracy of the objects not mentioned in captions using all kinds of tokens considered in our analysis (average final layer accuracy for primary object, which was 'Object in caption': CLS: 1 and avg\_token: 0.99, dropping to CLS: 0.87 and avg\_obj 0.93 when the object is 'not in caption'. In the secondary object, the drop is from 'Object in caption': CLS: 0.96 and avg\_obj: 0.99 to 'not in caption': CLS: 0.79 and avg\_obj: 0.82). This means the network pays more attention to certain objects, and its learning of discriminative features deteriorates for objects not mentioned in the captions. We note that the decrease in decoding accuracy is most pronounced for the CLS token, showing the direct effect of the downstream task on the representation.

Using CLIP in a multi-object setting, we further use this understanding to check for failure in zero-shot object detection tasks. We use the classification task setting of the secondary objects in the paired-object probe. Specifically, the object-specific representation (random\_token and avg\_obj token) is used to learn probes to classify the secondary objects in images. We also evaluate the CLIP zero-shot performance to detect objects in the image with the prompt "An image containing a 'category'", where the category was replaced with 4 object categories of the set. We finally evaluate the accuracy of detecting the objects. The results confirm the original CLIP paper by showing that learning a probe gives better accuracy than zero-shot accuracy in CLIP (Probe CLS: 0.65 vs Zero Shot: 57; see Fig. \ref{fig:4}.b). Further, we also note that the object-specific regions in the token space achieve far better representations than the CLS token (Random\_Obj: 0.76 and Avg\_Obj: 0.77). This emphasises that even though the CLS token is used to classify the objects in practice, comparing it to representations from the token space, we find that the object-specific regions perform much better in classification for CLIP. 

\subsection{Structured representations in the network are better to retain the representation of objects in the background}
\begin{figure*}
  \centering
   \includegraphics[width=1\linewidth]{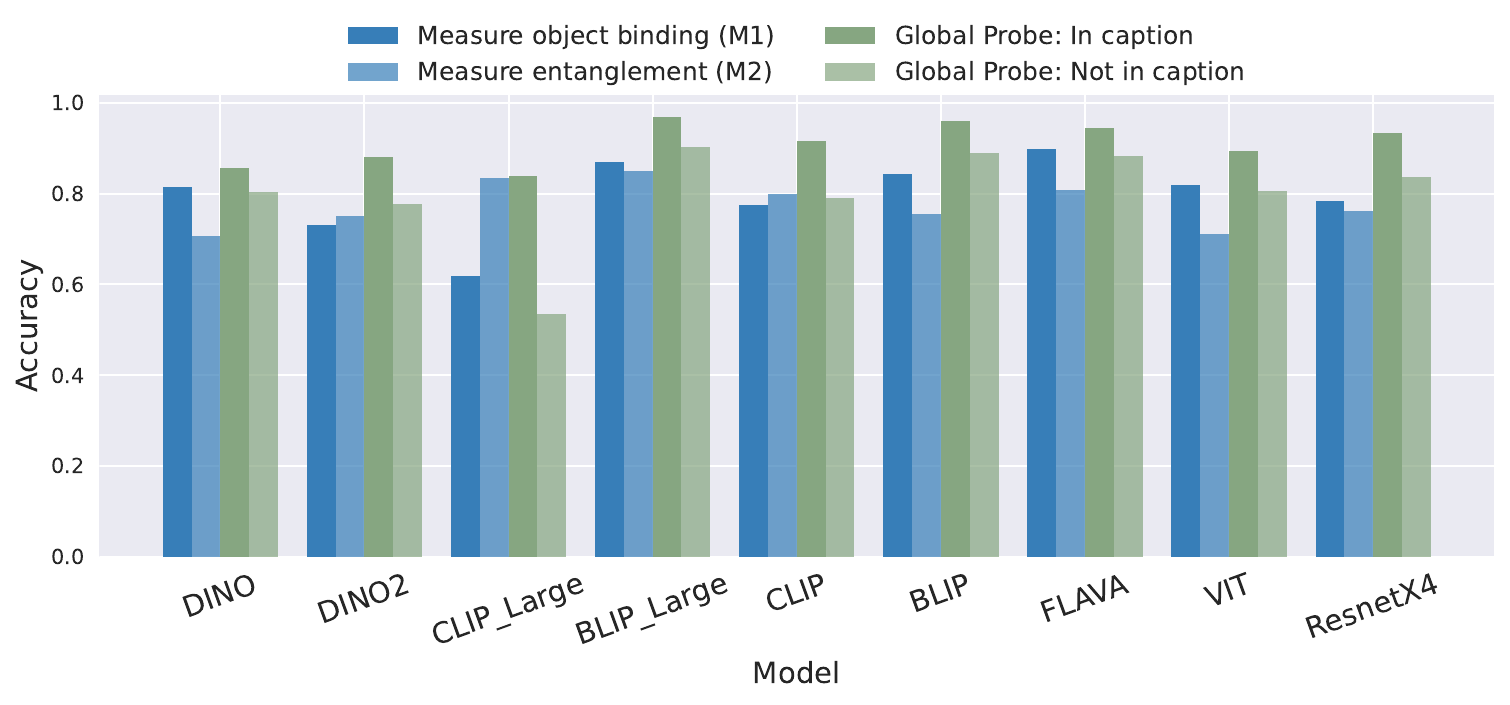}

   \caption{The figure shows both measures of modularity evaluated for all vision encoders. Along with this, the figure has the accuracy of global probing task for avg\_obj token evaluated for 'In caption': Objects mentioned in the caption and 'Not in caption': Objects not mentioned in the caption. The results for each network are from the layer with the best $M_1$ across the layers. We see how the first measure of modularity $M_1$ correlates highly with the accuracy of decoding background tokens across networks (Pearson's correlation coefficient = 0.94, p = 0.001). This also shows that the measures from the paired object task are highly predictive of general representation of objects in the network. The second measure of modularity (Measure of entanglement) $M_2$ correlates (Pearson's coefficient = 0.74, p = 0.02) with the drop in accuracy for objects 'not in the caption' compared to the objects mentioned 'in the caption'. Hence, more modular representations favour better representation of individual background objects.}
   \label{fig:5}
\end{figure*}

\begin{table}
\captionsetup{width=9cm}
\centering
\caption{Table showing the predicted layer with maximum object binding measure ($M_1$) and the actual layer with the best representation of background objects, i.e. Global Probe:'not in the caption' in the global decoding task.}
\begin{tabular}{ccccc}
\toprule
Model & Layer Predicted & Best layer\\
\midrule
DINO        & 8  & 7  &  &  \\
DINO2       & 10 & 10 &  &  \\
CLIP\_Large & 23 & 23 &  &  \\
BLIP\_Large & 20 & 21 &  &  \\
CLIP        & 10 & 9  &  &  \\
BLIP        & 9  & 9  &  &  \\
FLAVA       & 9 & 8  &  &  \\
VIT         & 7  & 10 &  &  \\
CLIP ResnetX4         & 3  & 3  &  & \\
\bottomrule
\end{tabular}
\label{tab:layer}
\end{table}

As discussed in the previous section, the object-specific tokens in the main token space of the network have better ability to retain and model object-specific information for most objects than the CLS token. This makes them especially useful for tasks requiring information about objects that may not be the scene's focus (according to the training distribution). However, a key question remains: how do we quantify this ability of object-specific tokens across layers and networks? Our proposed measures of structured representations address this by evaluating the degree to which networks model and maintain separate object-wise information in the token space. In Fig \ref{fig:5}, we show these measures across all networks using the Avg\_Obj tokens. Specifically, $M_1$ measures the decoding accuracy of the secondary object $O_s$ using secondary object token $T_s$. We find (see Fig \ref{fig:5}) that when networks exhibit strong modular representations of secondary objects, they also decode other 20 "background" objects more effectively if those objects are not mentioned in the COCO captions. Indeed, there is a strong Pearson's correlation of +0.94 with the object binding measure $M_1$ and a moderate inverse correlation of -0.53 (p-value = 0.13) with the entanglement measure $M_2$. The role of structured representations becomes clearer when comparing 'not in caption' objects to 'in caption' objects: the token entanglement measure $M_2$ correlates 0.74 (p-value = 0.02) with the drop in accuracy for 'not in caption' objects. This indicates that the more segregated the information in the tokens, the better a network's object-specific representation can be in equally retaining and decoding objects not mentioned in the caption (background objects). Hence, this shows that more modular representations in vision encoders, i.e. representation agreeing to our two properties of 1) binding and 2) segregation, are signs of better retention of object-specific representation of even less important 'objects not in the caption'. 

Our analysis also shows that in most encoders, the last layer is not the best choice for decoding the objects using the object-specific tokens from the networks; in models such as BLIP-L, object-wise representation degrades in the last two layers (0.87 in layer 20 to 0.73 in layer 23 for secondary objects), possibly due to the pressure of the objective function. In Table \ref{tab:layer}, we compare the estimated choice of the layer using the object binding measure $M_1$ and the layer with the best accuracy on background objects ('not mentioned in the caption'). Further, a lower entanglement score helped guide the choice when object binding measures are similar across layers. We see a good relative match between the estimated best layers using measures of modularity obtained using paired-object tasks and the best layers for the larger 20-class global decoding task. Hence, our measures of structured representations can be helpful in ascertaining which network and particular layers are useful to train decoders for tasks requiring information about background objects in multi-object scenarios. 

\section{Discussion}
In this work, we identify two properties required in representations of image encoders for them to be more structured: 1) The model should be able to bind information specific to particular objects in the image into specific
representation units, i.e. a limited number of tokens can represent the object better than its input
representation. 2) The model should segregate various objects of the input image in separate
sets of tokens, i.e., there should be object-wise
information disentanglement in the token space.

Most vision encoders fulfill the first criterion, as they aggregate information from the image into smaller representation spaces functional for the downstream task. In particular, the decoding accuracy for all objects is higher than that of their raw input representation. However, our results demonstrated that the bottleneck at the CLS token does not represent all objects well. 

We were interested in understanding whether separate representational units specialise in representing the image's constituent parts (objects). Our analysis shows that the object-specific (avg\_obj, random\_obj) tokens show decent decoding accuracy across the models. Each object-wise token’s accuracy for classifying itself is our estimate for fulfilling this criterion. This accuracy is better than for the CLS token for all models except CLIP-L.

For the second characteristic, we looked at the network’s capability to form disentangled representations of objects in the token space. We saw evidence that information about other objects also leaks into the object-wise representations. Ideally, each object representation can only decode that particular object, and other objects in the scene cannot be decoded by that representation. As the other objects can be decoded in our object representations (with accuracies of 0.89 by secondary object representations and 0.65 by primary object representations), their representations are already affected by the context, i.e. other objects. In addition, the higher the decoding accuracy of other objects in the scene, the higher the entanglement in the representation. 

Our comparative analysis of the BLIP model's performance showed a notable degradation in decoding efficacy across all representations when objects were omitted from its captions. We replicate this result using the object-wise tokens for all models examined in the global probe study. This observation underscores the significant influence of the model's downstream objective on its representational capabilities. Most research predominantly assesses these networks' capabilities based on their final outputs, which rely on representations like a CLS token (for example, in \cite{yuksekgonul2022and,lewis2022does}). Our findings show a loss in representing multiple objects while funnelling information into the CLS token. Further, the downstream objective's effect is the most extreme on the nature of the CLS representations. 

We show that this causes suboptimal performance of these networks in simple tasks such as identifying objects in the multi-object setting, especially for background objects. We show that the networks have some structured information in token space i.e. they preserve object-specific information localised in the token space. Our analysis finally shows that the measures of structured information in the token space of models correlate with the ability of the models to generalise on background objects using a 20-class experiment. This way, our measures of structured information evaluate the models' ability to represent individual objects in a multi-object scene. These measures show which model's representations and at what layer are most structured for representing individual objects separately. They achieve this by not combining object representations in the token space (other than CLS token) under the influence of objective functions that can lead to abstracting the background objects while modelling the image. Access to this structured representation helps with tasks requiring access to detailed representations of individual objects in the scene. 

The natural next step of this work is to verify the usefulness of the two representation measures on specific datasets (object categories) for the choice of the best layer and network on actual downstream tasks like semantic segmentation or object localisation. This way, one can estimate if the best-predicted layers for the networks give the best results, or in the case of the use of deep decoders for segmentation like in \cite{vobecky2025unsupervised,li2023mask,ayzenberg2024dinov2}, it will show how much less data is needed to train a decoder as opposed to using the default final layer (especially for good performance on less prominent background objects).

\section{Limitations}
 The probing and representation analysis methods we used have some limitations, and the results must be carefully interpreted. High decoding accuracy in a decoding task with two or four objects may be due to the easy decoding task (if the object classes are naturally distinct and it is only a 2/4 way classification). Further, the fact that one can classify a particular object from a few other classes may depend on representing a single or a few features distinguishing between the classes. This representation may not model many other aspects of the object. We controlled for this limitation by training global probes for 20-class classification using the same representations, showing reasonably high accuracy. Likewise, a lower decoding accuracy does not mean that the model does not have information about the particular object; the information is just not encoded with linear distinction at that layer/token. Overall, we are averaging over six object sets with many object categories. Hence, even though the exact numbers may not reflect the model's exact state, we can safely rely on the relative numbers to make inferences about the model and its representations.

\section{Conclusion}
In this work, we first formulated two characteristics for structured representations in an image encoder. We then analysed the representations of transformer-based image encoders in VIT, BLIP, CLIP, FLAVA, DINO, DINOv2 and CLIP (Resnet X4) image encoders. Through object decoding tasks, we create a view of the representation structure of these pre-trained models. We observed that object-specific areas hold the most discriminative features about the objects. Their discriminative ability decreased when the objects were not important for the downstream task but was still reasonably maintained. The token representations are not disentangled since they can decode other objects in the scene with accuracy far above random guesses. We found that the aggregate image representation, i.e. CLS token for transformers, does not represent all objects, and its discriminative ability degrades the most when the object is not useful for the trained downstream task. Our work thus characterises and measures the extent to which the representations of these image encoders are structured in the token space, that the individual objects are represented separately in the token space, but their representations are entangled. We also calculated the layerwise measures of representation and entanglement by object importance (estimated using captions), which provides insights into the failures and optimal adaptation of these networks to downstream tasks that utilise these individual object features.

\begin{credits}
\subsubsection{Supplementary Materials}

The code to obtain data splits from the COCO dataset, experimental setup, and data analysis can be found in the repository linked below. Further detailed plots for the paired object task showing the decoding accuracy of objects across the layers of the network have been given in the notebooks in the supplementary material for all models.

\url{https://github.com/tarunkhajuria42/Structured-representations}

\subsubsection{\ackname} We thank Raul Vicente, and Meelis Kull for the valuable discussions and inputs on the paper. This work was supported by the Estonian Research Council grant PSG728, the Estonian Centre of Excellence in Artificial Intelligence (EXAI), funded by the Estonian Ministry of Education and Research and the European Union’s Horizon 2020 Research and Innovation Programme under Grant Agreement No. 952060 (Trust AI).

\subsubsection{\discintname}
The authors declare that they have no competing interests.
\end{credits}
%
%
%
\bibliographystyle{splncs04}
\bibliography{sn-bibliography}

\begin{thebibliography}{10}
\providecommand{\url}[1]{\texttt{#1}}
\providecommand{\urlprefix}{URL }
\providecommand{\doi}[1]{https://doi.org/#1}

\bibitem{aflalo2022vl}
Aflalo, E., Du, M., Tseng, S.Y., Liu, Y., Wu, C., Duan, N., Lal, V.: Vl-interpret: An interactive visualization tool for interpreting vision-language transformers. In: Proceedings of the IEEE/CVF Conference on Computer Vision and Pattern Recognition. pp. 21406--21415 (2022)

\bibitem{alain2016understanding}
Alain, G., Bengio, Y.: Understanding intermediate layers using linear classifier probes. arXiv preprint arXiv:1610.01644  (2016)

\bibitem{antol2015vqa}
Antol, S., Agrawal, A., Lu, J., Mitchell, M., Batra, D., Zitnick, C.L., Parikh, D.: Vqa: Visual question answering. In: Proceedings of the IEEE international conference on computer vision. pp. 2425--2433 (2015)

\bibitem{ayzenberg2024dinov2}
Ayzenberg, L., Giryes, R., Greenspan, H.: Dinov2 based self supervised learning for few shot medical image segmentation. In: 2024 IEEE International Symposium on Biomedical Imaging (ISBI). pp.~1--5. IEEE (2024)

\bibitem{bengio2012representation}
Bengio, Y., Courville, A., Vincent, P.: Representation learning: A review and new perspectives. arxiv 2012. arXiv preprint arXiv:1206.5538  (2012)

\bibitem{bronstein2021geometric}
Bronstein, M.M., Bruna, J., Cohen, T., Veli{\v{c}}kovi{\'c}, P.: Geometric deep learning: Grids, groups, graphs, geodesics, and gauges. arXiv preprint arXiv:2104.13478  (2021)

\bibitem{cao2020behind}
Cao, J., Gan, Z., Cheng, Y., Yu, L., Chen, Y.C., Liu, J.: Behind the scene: Revealing the secrets of pre-trained vision-and-language models. In: Computer Vision--ECCV 2020: 16th European Conference, Glasgow, UK, August 23--28, 2020, Proceedings, Part VI 16. pp. 565--580. Springer (2020)

\bibitem{caron2021emerging}
Caron, M., Touvron, H., Misra, I., J{\'e}gou, H., Mairal, J., Bojanowski, P., Joulin, A.: Emerging properties in self-supervised vision transformers. In: Proceedings of the IEEE/CVF international conference on computer vision. pp. 9650--9660 (2021)

\bibitem{cordonnier2019relationship}
Cordonnier, J.B., Loukas, A., Jaggi, M.: On the relationship between self-attention and convolutional layers. arXiv preprint arXiv:1911.03584  (2019)

\bibitem{de2018clinically}
De~Fauw, J., Ledsam, J.R., Romera-Paredes, B., Nikolov, S., Tomasev, N., Blackwell, S., Askham, H., Glorot, X., O’Donoghue, B., Visentin, D., et~al.: Clinically applicable deep learning for diagnosis and referral in retinal disease. Nature medicine  \textbf{24}(9),  1342--1350 (2018)

\bibitem{fodor1998concepts}
Fodor, J.A.: Concepts: Where cognitive science went wrong. Oxford University Press (1998)

\bibitem{greff2020binding}
Greff, K., Van~Steenkiste, S., Schmidhuber, J.: On the binding problem in artificial neural networks. arXiv preprint arXiv:2012.05208  (2020)

\bibitem{hewitt2019designing}
Hewitt, J., Liang, P.: Designing and interpreting probes with control tasks. arXiv preprint arXiv:1909.03368  (2019)

\bibitem{johnson2017clevr}
Johnson, J., Hariharan, B., Van Der~Maaten, L., Fei-Fei, L., Lawrence~Zitnick, C., Girshick, R.: Clevr: A diagnostic dataset for compositional language and elementary visual reasoning. In: Proceedings of the IEEE conference on computer vision and pattern recognition. pp. 2901--2910 (2017)

\bibitem{koh2020concept}
Koh, P.W., Nguyen, T., Tang, Y.S., Mussmann, S., Pierson, E., Kim, B., Liang, P.: Concept bottleneck models. In: International conference on machine learning. pp. 5338--5348. PMLR (2020)

\bibitem{krishna2017visual}
Krishna, R., Zhu, Y., Groth, O., Johnson, J., Hata, K., Kravitz, J., Chen, S., Kalantidis, Y., Li, L.J., Shamma, D.A., et~al.: Visual genome: Connecting language and vision using crowdsourced dense image annotations. International journal of computer vision  \textbf{123},  32--73 (2017)

\bibitem{lake2023human}
Lake, B.M., Baroni, M.: Human-like systematic generalization through a meta-learning neural network. Nature pp.~1--7 (2023)

\bibitem{lepori2023break}
Lepori, M.A., Serre, T., Pavlick, E.: Break it down: evidence for structural compositionality in neural networks. arXiv preprint arXiv:2301.10884  (2023)

\bibitem{lewis2022does}
Lewis, M., Yu, Q., Merullo, J., Pavlick, E.: Does clip bind concepts? probing compositionality in large image models. arXiv preprint arXiv:2212.10537  (2022)

\bibitem{li2023mask}
Li, F., Zhang, H., Xu, H., Liu, S., Zhang, L., Ni, L.M., Shum, H.Y.: Mask dino: Towards a unified transformer-based framework for object detection and segmentation. In: Proceedings of the IEEE/CVF conference on computer vision and pattern recognition. pp. 3041--3050 (2023)

\bibitem{li2022blip}
Li, J., Li, D., Xiong, C., Hoi, S.: Blip: Bootstrapping language-image pre-training for unified vision-language understanding and generation. In: International Conference on Machine Learning. pp. 12888--12900. PMLR (2022)

\bibitem{lin2014microsoft}
Lin, T.Y., Maire, M., Belongie, S., Hays, J., Perona, P., Ramanan, D., Doll{\'a}r, P., Zitnick, C.L.: Microsoft coco: Common objects in context. In: Computer Vision--ECCV 2014: 13th European Conference, Zurich, Switzerland, September 6-12, 2014, Proceedings, Part V 13. pp. 740--755. Springer (2014)

\bibitem{lipton2018mythos}
Lipton, Z.C.: The mythos of model interpretability: In machine learning, the concept of interpretability is both important and slippery. Queue  \textbf{16}(3),  31--57 (2018)

\bibitem{locatello2020object}
Locatello, F., Weissenborn, D., Unterthiner, T., Mahendran, A., Heigold, G., Uszkoreit, J., Dosovitskiy, A., Kipf, T.: Object-centric learning with slot attention. Advances in Neural Information Processing Systems  \textbf{33},  11525--11538 (2020)

\bibitem{lovering2022unit}
Lovering, C., Pavlick, E.: Unit testing for concepts in neural networks. Transactions of the Association for Computational Linguistics  \textbf{10},  1193--1208 (2022)

\bibitem{oord2017neural}
Oord, A.v.d., Vinyals, O., Kavukcuoglu, K.: Neural discrete representation learning. arXiv preprint arXiv:1711.00937  (2017)

\bibitem{oquab2023dinov2}
Oquab, M., Darcet, T., Moutakanni, T., Vo, H., Szafraniec, M., Khalidov, V., Fernandez, P., Haziza, D., Massa, F., El-Nouby, A., et~al.: Dinov2: Learning robust visual features without supervision. arXiv preprint arXiv:2304.07193  (2023)

\bibitem{pavlick2023symbols}
Pavlick, E.: Symbols and grounding in large language models. Philosophical Transactions of the Royal Society A  \textbf{381}(2251),  20220041 (2023)

\bibitem{pedregosa2011scikit}
Pedregosa, F., Varoquaux, G., Gramfort, A., Michel, V., Thirion, B., Grisel, O., Blondel, M., Prettenhofer, P., Weiss, R., Dubourg, V., et~al.: Scikit-learn: Machine learning in python. the Journal of machine Learning research  \textbf{12},  2825--2830 (2011)

\bibitem{radford2021learning}
Radford, A., Kim, J.W., Hallacy, C., Ramesh, A., Goh, G., Agarwal, S., Sastry, G., Askell, A., Mishkin, P., Clark, J., et~al.: Learning transferable visual models from natural language supervision. In: International conference on machine learning. pp. 8748--8763. PMLR (2021)

\bibitem{radulescu2021human}
Radulescu, A., Shin, Y.S., Niv, Y.: Human representation learning. Annual Review of Neuroscience  \textbf{44}(1),  253--273 (2021)

\bibitem{raghu2021vision}
Raghu, M., Unterthiner, T., Kornblith, S., Zhang, C., Dosovitskiy, A.: Do vision transformers see like convolutional neural networks? Advances in Neural Information Processing Systems  \textbf{34},  12116--12128 (2021)

\bibitem{rudin2019stop}
Rudin, C.: Stop explaining black box machine learning models for high stakes decisions and use interpretable models instead. Nature machine intelligence  \textbf{1}(5),  206--215 (2019)

\bibitem{shridhar2021cliport}
Shridhar, M., Manuelli, L., Fox, D.: Cliport: What and where pathways for robotic 569 manipulation. In: Conference on Robot Learning. pp. 894--906 (2021)

\bibitem{singh2022flava}
Singh, A., Hu, R., Goswami, V., Couairon, G., Galuba, W., Rohrbach, M., Kiela, D.: Flava: A foundational language and vision alignment model. In: Proceedings of the IEEE/CVF Conference on Computer Vision and Pattern Recognition. pp. 15638--15650 (2022)

\bibitem{trauble2023discrete}
Tr{\"a}uble, F., Goyal, A., Rahaman, N., Mozer, M.C., Kawaguchi, K., Bengio, Y., Sch{\"o}lkopf, B.: Discrete key-value bottleneck. In: International Conference on Machine Learning. pp. 34431--34455. PMLR (2023)

\bibitem{vobecky2025unsupervised}
Vobecky, A., Hurych, D., Sim{\'e}oni, O., Gidaris, S., Bursuc, A., P{\'e}rez, P., Sivic, J.: Unsupervised semantic segmentation of urban scenes via cross-modal distillation. International Journal of Computer Vision pp. 1--23 (2025)

\bibitem{yi2018neural}
Yi, K., Wu, J., Gan, C., Torralba, A., Kohli, P., Tenenbaum, J.: Neural-symbolic vqa: Disentangling reasoning from vision and language understanding. Advances in neural information processing systems  \textbf{31} (2018)

\bibitem{yuksekgonul2022and}
Yuksekgonul, M., Bianchi, F., Kalluri, P., Jurafsky, D., Zou, J.: When and why vision-language models behave like bags-of-words, and what to do about it? In: The Eleventh International Conference on Learning Representations (2022)

\bibitem{yun2022vision}
Yun, T., Bhalla, U., Pavlick, E., Sun, C.: Do vision-language pretrained models learn composable primitive concepts? arXiv preprint arXiv:2203.17271  (2022)

\end{thebibliography}
%




 \end{document}